\documentclass[10pt, a4paper]{article}
\usepackage{swedish_bert}
\usepackage{multibib}
\newcites{languageresource}{Language Resources}
\usepackage{graphicx}
\usepackage{tabularx}
\usepackage{soul}

\usepackage{epstopdf}
\usepackage[utf8]{inputenc}
\usepackage{MnSymbol,wasysym}
\usepackage{hyperref}
\usepackage{xstring}

\usepackage{color}

\title{ \vspace*{.5\baselineskip} \textbf{Playing with Words at the National Library of Sweden - \\Making a Swedish BERT}}

\name{Martin Malmsten, Love Börjeson, Chris Haffenden}

\address{KBLab, National Library of Sweden \\
         Humlegården, Stockholm \\
         www.kb.se/kb-labb\\
         \{martin.malmsten, love.borjeson, chris.haffenden\}@kb.se\\}

\abstract{
This paper introduces the Swedish BERT (“KB-BERT”) developed by the KBLab for data-driven research at the National Library of Sweden (KB). Building on recent efforts to create transformer-based BERT models for languages other than English, we explain how we used KB’s collections to create and train a new language-specific BERT model for Swedish. We also present the results of our model in comparison with existing models—chiefly that produced by the Swedish Public Employment Service, Arbetsförmedlingen, and Google’s multilingual M-BERT—where we demonstrate that KB-BERT outperforms these in a range of NLP tasks from named entity recognition (NER) to part-of-speech tagging (POS). Our discussion highlights the difficulties that continue to exist given the lack of training data and testbeds for smaller languages like Swedish. We release our model for further exploration and research here: https://github.com/Kungbib/swedish-bert-models. \\ \newline  }

\begin{document}

\maketitleabstract

\section{Introduction}

Recent advances in neural network architectures have enabled significant new developments within natural language processing. With the appearance of the first BERT model—Bidirectional Encoder Representations from Transformers—Google’s researchers could point towards state-of-the-art performance across a variety of NLP tasks in English \cite{devlin_bert_2019}. This was followed by the development of M-BERT, a multilingual model capable of generalizing across languages \cite{pires_how_2019,wu_beto_2019}, which in turn prompted various monolingual models beyond English—i.e. CamemBERT and FinBERT—that have been shown to outperform M-BERT for deep transfer learning of these languages \cite{martin_camembert_2020,virtanen_multilingual_2019,ronnqvist_is_2019}. While demonstrating impressive performance on a number of tasks, these latter models have needed to overcome the challenges of a lack of data and testbeds for languages other than English. 

Here we present the new Swedish model we have developed at the National Library of Sweden (KB): KB-BERT. As with the likes of FinBERT, this has meant finding solutions to the lack of data for a lower-resourced language. We elaborate upon how our work at the recently-established KBLab for data-driven research has allowed us to address this problem — at least in part (see discussion below). Technical issues of access to data converge with wider questions about the representativeness of language models: we have deliberately used parts of the library’s collections with significant elements of colloquial language, including the newspaper archive and government reports—both containing large amounts of direct reported speech from a wide spectrum of social actors—to create a BERT trained on a broad range of Swedish. We have also included historical materials from the newspaper archive dating back to 1945 to capture something approximate to the living language of the national community \cite{benedict_imagined_2016,ahearn_living_2017}. There has thus been a significant democratizing of data in the training of KB-BERT.

This paper makes several contributions. Firstly, we explain how we used KB’s collections to create the training data necessary for the making of our Swedish BERT. Secondly, we describe and show the results from our testing process, where we compare the performance of KB-BERT for various NLP tasks in Swedish with that of existing models. Thirdly, we evaluate this performance and highlight the particular steps for future research that we think this raises.  


\section{ Corpus disposition }

The main challenge that we faced in developing KB-BERT—apart from the amount of computational resources needed—was in compiling and cleaning a sufficiently large and diverse corpus of Swedish text, cf. \cite{linder_automatic_2020}. Fortunately, we have been able to draw upon KB’s vast textual resources, which result from both large-scale digitization projects and the library’s role as the national archive for electronic and traditional legal deposits.

To limit the corpus solely to modern Swedish, we have selected only resources created from the 1940’s until late 2019. More precisely, we have included material from the following categories:

\begin{itemize}
    \item{Digitized newspapers—text extracted from OCR’d newspapers from KB’s archives. This newspaper corpus contains a wide variety of text such as quotes, speech, articles, short stories and adverts.}
    \item{Official Reports of the Swedish Government—the collection of government publications (Statens offentliga utredningar or SOU) contain around 5,000 volumes, ranging from 1940 to today. These texts are highly vetted before publication and represent a large amount of correctly written Swedish. The sentences tend to be longer than in other texts.}
    \item{Legal e-deposits—resources deposited at KB under the e-deposit law (2012: 492). These include reports from Swedish authorities, e-books from publishers and e-magazines.}
    \item{Social media—comments collected from internet forums. This is by far the most challenging corpus in terms of sentence structure and generally creative use of the Swedish language. Punctuation is, for example, apparently optional in such texts.}
    \item{Swedish Wikipedia—all Wikipedia articles in Swedish.}
\end{itemize}

\begin{table}[!h]
\small
\begin{center}
\begin{tabularx}
      {\columnwidth}{X r r r}
      \textbf{Corpus}          & \textbf{Words}  & \textbf{Sentences} & \textbf{Size} \\ \hline
      allnews.txt     & 2 997M & 226M      & 16 783MB  \\ 
      sou.txt         & 117M   & 7M        & 834MB     \\
      e-deposit.txt   & 62M    & 3.5M      & 400MB     \\
      social.txt      & 31M    & 2.2M      & 163MB     \\
      sw-wiki.txt     & 29M    & 2.1M      & 161MB     \\ \hline
      Total           & 3 497M & 260M      & 18 341MB  \\
\end{tabularx}
\caption{Size of cleaned corpora}
\end{center}
\end{table}

As can be seen in the table above, there is an obvious skew towards newspaper text. On the other hand, the \textit{allnews.txt} corpus is the most diverse of the corpora we have included. It is also our understanding that the amount of a certain type of text is not proportional to the model’s subsequent understanding of that type of text. In other words, incorporating smaller corpora of different types of text, rather than simply adding more text of an existing type, is still beneficial not only to the model’s representativity but also its capacity and robustness as a whole.
 
The one notable omission is the comparatively low amount of fiction included in the total corpus. This results from the fact that there is only a small number of e-books in the combined corpus. The newspaper corpus has a certain level of fictional texts, but the extent is unknown and it is mostly short stories rather than full books. The legal e-deposit corpus does, however, include a few hundred works of fiction.

\section{Text wrangling}
The text in its raw form needed extensive cleaning and filtering before it could be used to create pre-training data. Text preparation was carried out using a combination of available and custom-made tools to deal with discrepancies between and within corpora, as we describe below.

\subsection{OCR quality fix}
One major hurdle has been the presence of a few very specific OCR errors,  which most likely stem from an overly aggressive use of word lists, or a faulty word list, in combination with the OCR engine confusing the letter m with the letter r followed by n—and vice versa. This is especially problematic for the word om, meaning about in Swedish, being OCRed as orn (which is not a Swedish word) and then replaced by örn (meaning eagle in Swedish). Since it is a very common word showing up in ca 82\% of all newspaper pages, where more than half of all occurrences were incorrectly OCRed on this point, the resulting model would have ended up with a faulty understanding of Swedish.

This also became problematic with common endings in Swedish such as “arna”/”erna”, which would be OCRed as “ama” / “ema”. For example, “tanterna” [ “the old ladies” ] becomes “tantema” (not a Swedish word). To correct this we developed a fix to identify candidates where m had been replaced with rn and vice versa, by checking both versions against The Language Bank in Gothenburg’s (Språkbanken) morphological thesaurus: Saldo \cite{borin_saldo_2013}. If the original word did not exist in Saldo, but the transformed one did, the word was added as a candidate in a mapping file. 

In the end, over 30,000 such candidates were identified—the majority appearing only a few times each— and corrected. In the case of om -- örn the word was simply replaced, essentially deleting mentions of actual eagles from the Swedish newspaper corpus. In total, 13 million occurrences were corrected.

\subsection{Sentence- and paragraph splitting}
The pre-training data creation process ultimately demands “paragraphs”. This can mean quite different things, depending on the type of text at hand. In newspaper text it is usually a block of text spanning part of a page, while in monographs it is not uncommon for a paragraph to cover the whole page. In newspapers it is also not uncommon for single words or sentences to show up as paragraphs in the OCR. These are not valuable since they have little or no context. To reduce noise, we therefore set a somewhat arbitrary lower limit for a paragraph at ten sentences. This meant that we discarded quite a large amount of paragraphs that were deemed too short. Sentences at the beginning and end of paragraphs that did not seem to be whole sentences were also discarded.

However, the major obstacle to creating paragraphs that we faced was in splitting text into sentences. In the end, we used a custom script together with spaCy\cite{spacy2} to try and recreate sentences from a stream of text. This consisted of a pre-processing step to deal with hyphenation—which is very common since Swedish has fairly long composite words and columns of text can be narrow—followed by Spacy’s sentence splitter, and lastly a post-process to deal with special cases not covered by spaCy. It is worth noting that SpaCy does not yet have a statistical model for Swedish, which is partly why so much custom scripting was required. 

\subsection{Emojis}
A special case is that of the emojis that are occasionally used as punctuation in texts on social media. For example, the phrase “Hej \smiley Vad heter du?” [“Hi \smiley What’s your name?”]  is two sentences since the U+1F600 GRINNING FACE works as end-of-sentence punctuation in this context. On the other hand, “Menvafan \frownie \frownie \frownie varför gjorde du så!?” [“What the hell \frownie\frownie \frownie why did you do that?”]  is one sentence even though it contains three emojis. We felt it important to retain emojis both in the tokenization and in their use as sentence delimiters since they are prevalent in some types of discourse, and if the model was to have any chance of being used to analyze and/or understand conversations in social media, this was necessary.

\subsection{Tokenization}
Once the text was cleaned and split into paragraphs of sentences, we used the SentencePiece \cite{DBLP:journals/corr/abs-1808-06226} library to create a tokenizer file and then manually added emojis and skin color modifiers that had been filtered out for some reason (Kudo and Richardson, 2018). Following the lead of both the German \cite{branden_open_2019} and Finnish BERT \cite{virtanen_multilingual_2019} projects, we chose a dictionary size of approximately 50,000 tokens, with the rationale that Swedish contains many compound words.

\section{Pre-training}
We used the code and instructions provided in Google’s BERT repo to create pre-training data and subsequently pre-train the model \cite{devlin_bert_2019}. We trained for one million steps with a max sequence length of 128 and a batch size of 512. Then we trained for 100,000 steps with a msl of 512 and batch size of 128. We continued training for approximately another million steps with msl of 128 and batch size of 512 to see if this would yield better results downstream.

Model pretraining was made partly in-house at the KBLab and partly, for unproblematic material without active copyright, with generous support of Cloud TPUs from Google's TensorFlow Research Cloud (TFRC).


\section{Results and performance comparison}
To evaluate the performance of the model and compare it with others we used NER- and POS-tagging as a downstream task. These are both token classification task that require some language understanding to perform well.

\subsection{NER tagging}
For this task, we used the Stockholm-Umeå Corpus 3.0 \citelanguageresource{suc3} dataset hosted by The Language Bank in Gothenburg (Språkbanken) converted to be used by a modified version of the \textit{run\_ner.py} program included in the Huggingface\cite{Wolf2019HuggingFacesTS} framework. The file was split into a training, test and evaluation set, with 70\% used for training, and 20\% and 10\% for test and evaluation respectively.

The results show that our model, KB-BERT, outperforms existing BERTs either trained for multilingual understanding by Google or specifically for Swedish by Arbetsförmedlingen. Results for HFST-SweNER \cite{kokkinakis_hfst-swener_2014} are also included in the table below, but a difference in evaluation method means that the numbers are not necessarily comparable. 

\begin{table}[h!]
\small
\begin{center}
\tabcolsep=0.11cm
\begin{tabularx}
      {\columnwidth}{X r r r r}
      \textbf{Tag} & \textbf{AF-AI} & \textbf{M-BERT}  & \textbf{KB-BERT} & \textbf{HFST*} \\ \hline
      PER     & 0.913 & 0.945 & \textbf{0.961} & 0.913 \\ 
      ORG     & 0.780 & 0.834 & \textbf{0.884} & 0.534 \\ 
      LOC     & 0.913 & 0.942 & \textbf{0.958} & 0.780 \\ 
      TME     & 0.655 & 0.888 & \textbf{0.906} & -- \\ 
      MSR     & 0.828 & 0.853 & \textbf{0.890} & -- \\
      WRK     & 0.596 & 0.631 & \textbf{0.720} & 0.275 \\
      EVN     & 0.716 & 0.792 & \textbf{0.834} & 0.513 \\ 
      OBJ     & 0.710 & 0.761 & \textbf{0.770} & 0.437 \\ \hline
      AVG     & 0.876 & 0.906 & \textbf{0.927} & -- \\
\end{tabularx}
\caption{FB1 for NER tags}
\end{center}
\end{table}

The model from Arbetsförmedlingen is trained using only a fraction of the data used for KB-BERT, which shows in its limited performance.

\subsection{Part-of-spech tagging}
Using the SUC dataset we evaluated Part-of-speech tagging the same way as NER. The results show a much smaller improvement, less than 1\%, over M-BERT compared to NER tagging. This is probably due to F1 being very high to begin with since a lot of words have a static mapping to a corresponding POS tag.

\begin{table}[h!]
\footnotesize
\begin{center}
\tabcolsep=0.11cm
\begin{tabularx}
      {\columnwidth}{X r r ||X r r}
      \textbf{Tag} & \textbf{KB-BERT} & \textbf{BERT-M} & \textbf{Tag} & \textbf{KB-BERT} & \textbf{BERT-M} \\ \hline
      AB    & 0.9495 & 0.9421 & PC & 0.9526 & 0.9327 \\
      DT    & 0.9905 & 0.9876 & PL & 0.9131 & 0.8917\\
      HA    & 0.9274 & 0.9249 & PM & 0.9785 & 0.9689 \\
      HD    & 1.0000 & 1.0000 & PN & 0.9899 & 0.9874 \\
      HP    & 0.9727 & 0.9700 & PP & 0.9923 & 0.9889\\
      HS    & 1.0000 & 1.0000 & PS & 0.9976 & 0.9961 \\
      IE    & 0.9992 & 0.9984 & RG & 0.9696 & 0.9664 \\
      IN    & 0.9574 & 0.9355 & RO & 0.9610 & 0.9554 \\
      JJ    & 0.9687 & 0.9579 & SN & 0.9844 & 0.9804 \\
      KN    & 0.9876 & 0.9860 & UO & 0.8094 & 0.7197\\
      NN    & 0.9941 & 0.9896 & VB & 0.9987 & 0.9954 \\ \hline
      AVG   & \textbf{0.9863} & 0.9819 & & & \\ 
\end{tabularx}
\caption{FB1 for POS tags}
\end{center}
\end{table}

\subsection{Downstream tasks evaluation during pre-training}
To get a sense of when to stop pre-training, evaluation of the NER task was continually made at certain checkpoints as a measure of model performance. Results show a fairly slow rise after an initial jump somewhere around ten thousand steps with only marginal improvements after a few hundred thousand steps. This is in line with the results from Deepset’s German BERT(Chan et al., 2019).

\begin{table}[h!]
\small
\begin{center}
\tabcolsep=0.11cm
\begin{tabularx}
      {\columnwidth}{X r r r r r r r}
      \textbf{Task} & \textbf{10k} & \textbf{50k}  & \textbf{150k} & \textbf{350k} & \textbf{700k} & \textbf{1M} & \textbf{2M} \\ \hline
      NER (F1) & 0.8687 & 0.912 & 0.918 & 0.926 & 0.925 & 0.923 & 0.927 \\ 
\end{tabularx}
\caption{FB1 for NER during pre-training}
\end{center}
\end{table}

\subsection{Results for harder NER tasks}
While our KB-BERT scores objectively high and outperforms existing models, some specific tests designed to stress language understanding in downstream tasks work only intermittently. For example, the latter part of the sentence “Pelle och Kalle startar företaget Pelle och Kalle” [“Pelle and Kalle start the company Pelle and Kalle”] gets correctly tagged as ORG in some runs but not in others. This does not seem to stabilize with additional pre-training above 1-2 million steps. However, there does seem to be a difference for more complex NER tasks in the range between 350k and 1M pre-training steps. Perhaps we are pushing the boundaries of the model, or perhaps more fine-tuning data is needed. Further research is required to clarify this.

\section{Conclusion}
In this paper, we have described how we used KB’s considerable textual resources to train a new Swedish BERT model. Looking at the performance of KB-BERT, we have demonstrated that it outperforms both Google’s multilingual model, M-BERT, and existing Swedish models, such as Arbetsförmedlingen’s, across a range of NLP tasks. 

What conclusions can be drawn from this? Firstly, and most obviously, it highlights the importance of data volume: the more data, the better the performance. That a BERT model trained specifically for Swedish outperforms a multilingual model is hardly surprising; this has been shown with comparable results in other languages such as German and Finnish. However, the clear improvement in language understanding of KB-BERT in relation to other Swedish models is interesting, since it demonstrates the significance of the large amount of data required if a model is to perform well. 

Secondly, this study suggests the clear value of including a broad range of language types in creating a truly robust model. Perhaps most surprising here was that the use of apparently messy data, such as the formally unorthodox and colloquial language used on social media that incorporates emojis, seems to result in a BERT-model with a better capacity for understanding many different types of language (not just specifically that used on social media). In general terms, we believe that this supports our contention that a democratically-inclined model trained on data derived from many social domains leads to improved performance on NLP tasks. We intend to develop our analysis of this issue further in future research at KBLab. 

Thirdly, our work in this project has highlighted the continued problem of a lack of testbeds for the Swedish language. In contrast to well-resourced languages such as English, this lack presents particular problems for the future development and testing of new Swedish language models. To help address this issue, we are going to be participating in a new project aiming to develop new and improved testbeds in Swedish over the coming year, in collaboration with researchers at the Swedish Research Institute (RISE), the Swedish Language Bank at Gothenburg University, and AI Innovation of Sweden (“SuperLim: en svensk testmängd för språkmodeller”, funded by Vinnova, 2020-2021). Our hope is that this will lead to significantly improved possibilities for the future evaluation of Swedish models. 

\section{Acknowledgements}

\begin{itemize}
    \item{National Library of Sweden / KBLab}
    \item{Resources from Stockholm University, Umeå University and Swedish Language Bank at Gothenburg University were used when fine-tuning KB-BERT for NER.}
    \item{Research supported with Cloud TPUs from Google's TensorFlow Research Cloud (TFRC)}
    \item{Models are hosted on S3 by Huggingface} 
\end{itemize}

\section{Bibliographical References}\label{reference}

\bibliographystyle{swedish_bert}
\bibliography{swedish_bert}

\section{Language Resource References}
\label{lr:ref}
\bibliographystylelanguageresource{swedish_bert}
\bibliographylanguageresource{languageresource}

\end{document}